\def\paperID{0000}
\def\confName{CVPR}
\def\confYear{2025}

\def\paperTitle{IGR: Improving Diffusion Model for Garment Restoration from Person Image}

\def\authorBlock{
    Le Shen \qquad
    Rong Huang \qquad
    Zhijie Wang \\
    Donghua University, Shanghai, China \\
    {\tt\small \{le.shen\}@mail.dhu.edu.cn \{rong.huang, wangzj\}@dhu.edu.cn}
}

\newif\ifreview 
\newif\ifarxiv \newcommand{\arxiv}{\arxivtrue}
\newif\ifcamera 
\newif\ifrebuttal 

\arxiv
\pdfoutput=1
\documentclass[10pt,twocolumn,letterpaper]{article}
\ifreview \usepackage[review]{cvpr} \fi
\ifarxiv \usepackage[pagenumbers]{cvpr} \fi
\ifrebuttal \usepackage[rebuttal]{cvpr} \fi
\ifcamera \usepackage{cvpr} \fi

\usepackage{graphicx}	
\usepackage{amsmath}	
\usepackage{amssymb}	
\usepackage{booktabs}
\usepackage{times}
\usepackage{microtype}
\usepackage{epsfig}
\usepackage{caption}
\usepackage{float}
\usepackage{placeins}
\usepackage{color, colortbl}
\usepackage{stfloats}
\usepackage{enumitem}
\usepackage{tabularx}
\usepackage{xstring}
\usepackage{multirow}
\usepackage{xspace}
\usepackage{url}
\usepackage{subcaption}
\usepackage{xcolor}
\usepackage[hang,flushmargin]{footmisc}

\ifcamera \usepackage[accsupp]{axessibility} \fi

% ...

\ifarxiv  \fi

\newcommand{\R}[1]{{%
    \textbf{%
        \ifstrequal{#1}{1}{\textcolor{red}{R#1}}{%
        \ifstrequal{#1}{2}{\textcolor{blue}{R#1}}{%
        \ifstrequal{#1}{3}{\textcolor{magenta}{R#1}}{%
        \ifstrequal{#1}{4}{\textcolor{teal}{R#1}}{%
                           \textcolor{cyan}{R#1}%
        }}}}%
    }%
}}

\usepackage{xr-hyper}

\makeatletter
\newcommand*{\addFileDependency}[1]{
  \typeout{(#1)}
  \@addtofilelist{#1}
  \IfFileExists{#1}{}{\typeout{No file #1.}}
}

\makeatother

\definecolor{cvprblue}{rgb}{0.21,0.49,0.74}
\usepackage[pagebackref,breaklinks,colorlinks,allcolors=cvprblue]{hyperref}
\usepackage[capitalize]{cleveref}
\crefname{section}{Sec.}{Secs.}
\crefname{table}{Table}{Tables}
\crefname{figure}{Fig.}{Figs.}

\ifarxiv \crefname{appendix}{App.}{Apps.}
\else \crefname{appendix}{Suppl.}{Suppls.} \fi

\begin{document}

\title{\paperTitle}
\author{\authorBlock}

\twocolumn[{%
\renewcommand\twocolumn[1][]{#1}%
\maketitle
\begin{center}
    \centering
    \captionsetup{type=figure}
        \includegraphics[width=0.99\linewidth]{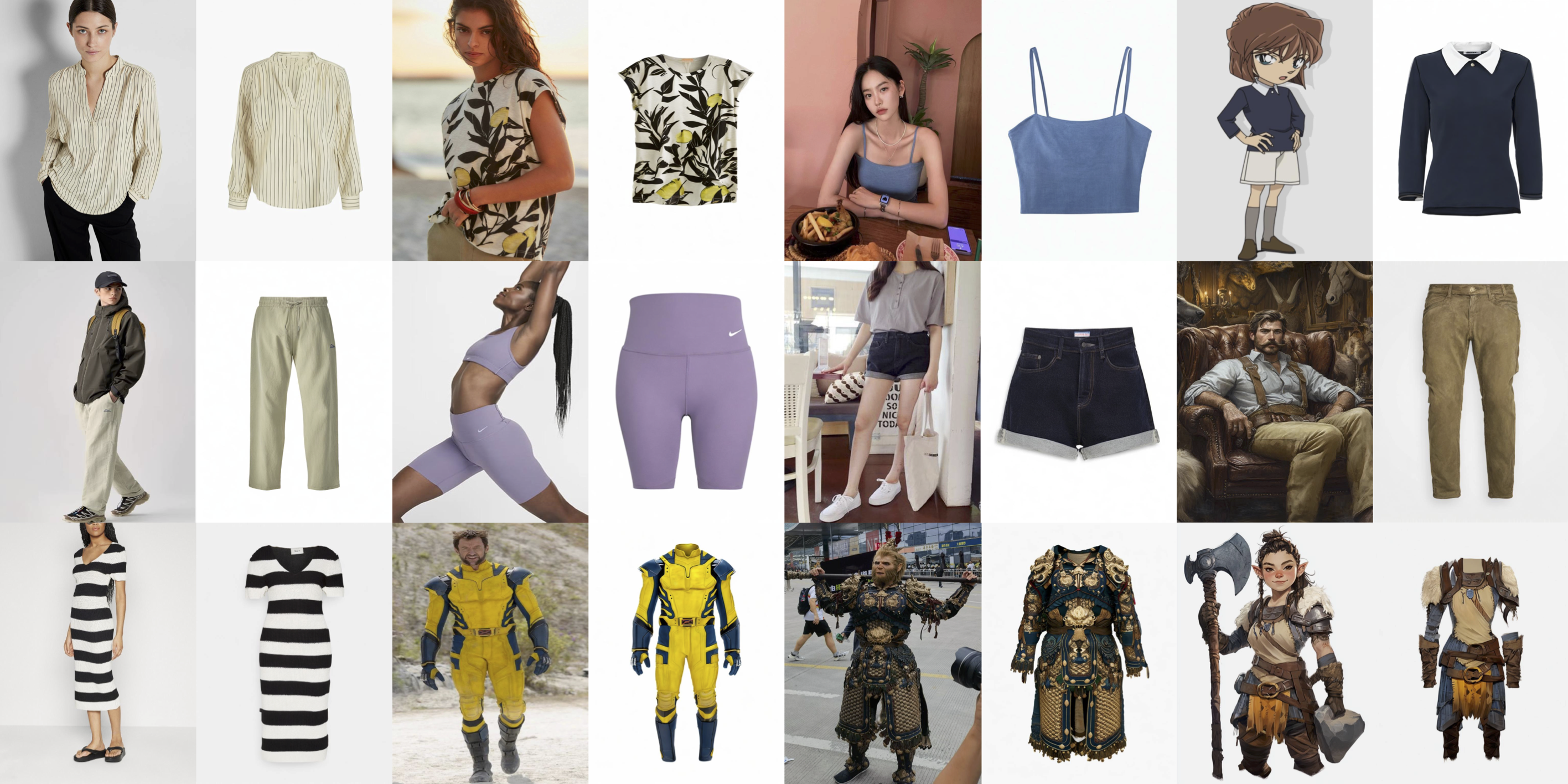}
    \captionof{figure}{Several pairs of reference person image and our restored standard garment. From top to bottom, the pairs demonstrate the restoration of upper, lower, and full-body garments.}
    \label{fig:showresult}
\end{center}%
}]

\begin{abstract}
Garment restoration, the inverse of virtual try-on task, focuses on restoring standard garment from a person image, requiring accurate capture of garment details. However, existing methods often fail to preserve the identity of the garment or rely on complex processes. To address these limitations, we propose an improved diffusion model for restoring authentic garments. Our approach employs two garment extractors to independently capture low-level features and high-level semantics from the person image. Leveraging a pretrained latent diffusion model, these features are integrated into the denoising process through garment fusion blocks, which combine self-attention and cross-attention layers to align the restored garment with the person image. Furthermore, a coarse-to-fine training strategy is introduced to enhance the fidelity and authenticity of the generated garments. Experimental results demonstrate that our model effectively preserves garment identity and generates high-quality restorations, even in challenging scenarios such as complex garments or those with occlusions.
\end{abstract}

\section{Introduction}
\label{sec:intro}

\begin{figure}[ht]
    \centering
    \includegraphics[width=1\linewidth]{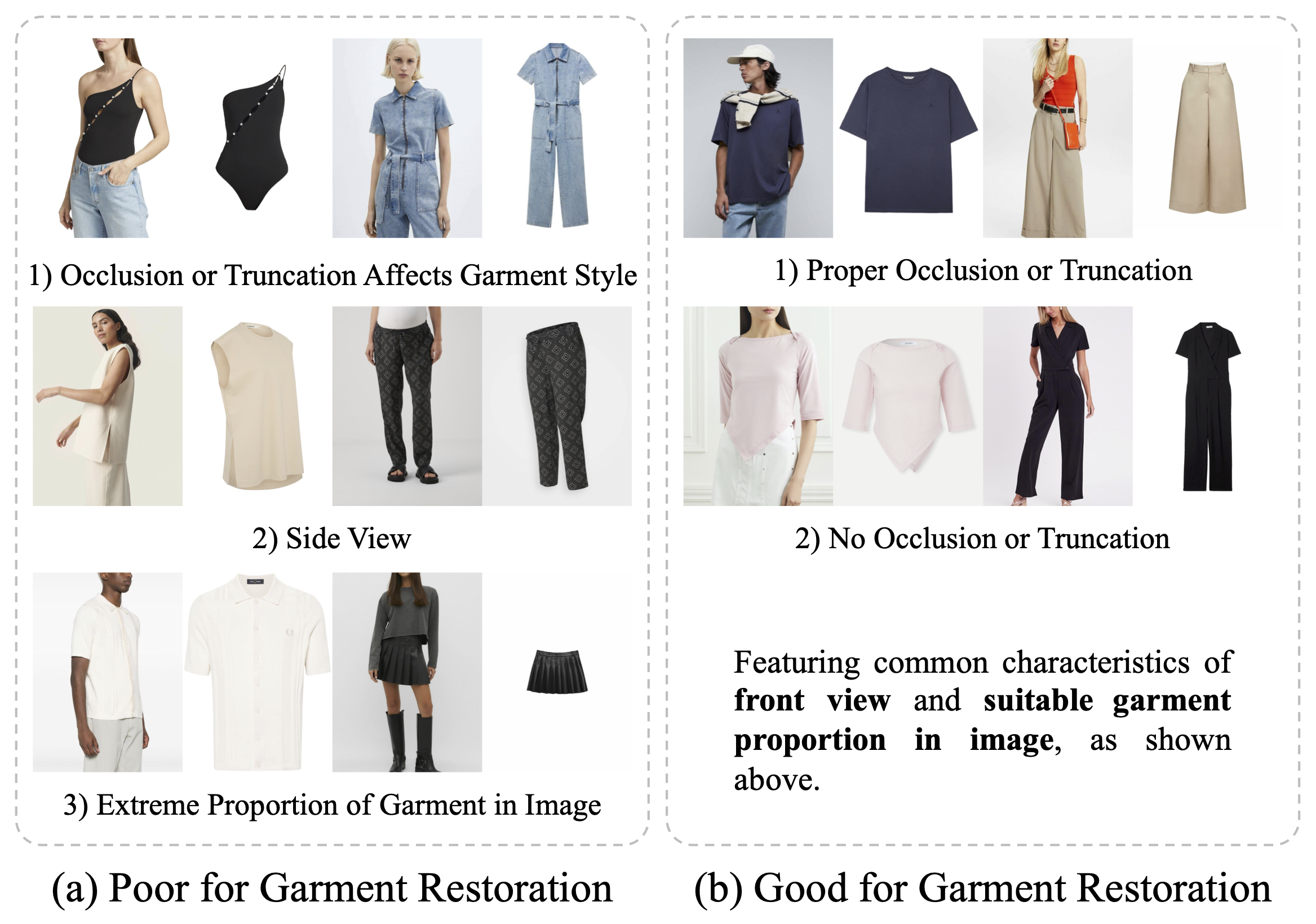}
    \caption{Training data for GarmRe. The left part displays three types of data pairs that may lead to poor training. The right part showcases the data pairs perfectly matched for training.}
    \label{fig:fig2_dataset}
\end{figure} 

With the rapid development of diffusion models~\cite{ho2020denoising, song2020denoising, rombach2022high}, great breakthroughs have been made in garment related downstream tasks~\cite{xu2024tunnel,karras2024fashion, niu2024anydesign, zhang2025garmentaligner}, especially in the field of virtual try-on (VITON) task~\cite{choi2021viton, kim2024stableviton, cui2024street, choi2024improving, xu2024ootdiffusion}. The realistic generative results have made VITON increasingly valuable in the fashion e-commerce field~\cite{zalando2024}, enabling users to preview try-on outcomes. However, users sometimes wish to wear the exact garment shown on a reference person. A straightforward approach is to use powerful segmentation methods~\cite{ravi2024sam, kirillov2023segment} to extract garment from the reference and apply it to virtual try-on. Unfortunately, most methods fail to capture complete garment details and often produce unrealistic try-on images. A more effective solution~\cite{zeng2020TileGAN, velioglu2024tryoffdiff,xarchakos2024tryoffanyonetiledclothgeneration, tan2024ragdiffusion} is to first restore the standard garment with clean background from the reference person, namely, garment restoration (GarmRe) and then use it for the try-on process. The main challenge of this task is to accurately capture the garment details from the person image and produce a standard garment that maintains the same identity, without missing characteristics or introducing artifacts.

Despite its potential applications in VITON systems, this garment restoration task remains largely under-explored by researchers~\cite{zeng2020TileGAN, velioglu2024tryoffdiff,xarchakos2024tryoffanyonetiledclothgeneration, tan2024ragdiffusion}.TileGAN~\cite{zeng2020TileGAN} is the first GAN-based~\cite{goodfellow2014generative, isola2017image} model for garment restoration, introduced in 2020. However, its generated results are limited by the capabilities of its foundational model, and the lack of application scenarios at the time leads to no further research. With the advancement in generative models~\cite{ho2020denoising, song2020denoising, rombach2022high} and the increasing demand for standard garments in VITON task, new approaches have recently emerged. TryOffDiff~\cite{velioglu2024tryoffdiff} and TryOffAnyone~\cite{xarchakos2024tryoffanyonetiledclothgeneration} leverage pretrained latent diffusion models (LDMs) as their foundation, achieving significant improvements thanks to LDMs' strong generative capabilities. TryOffDiff uses SigLIP~\cite{zhai2023sigmoid} as image encoder to extract garment features from person image, while TryOffAnyone directly concatenates person image with the denoiser inputs without additional model structure. Despite these improvements, both methods struggle to capture garment features effectively, resulting in generated garment that lack details. RAGDiffusion~\cite{tan2024ragdiffusion} incorporates retrieval-augmented techniques~\cite{borgeaud2022improving,ram2023context} to enhance garment generation. However, it relies on four separate processes to achieve its results, making it both time-consuming and overly complex.

In this paper, we propose IGR, a Improved diffusion-based Garment Restoration method, to generating true-to-life garment aligned with user-provided person image. We base our approach on Stable Diffusion~\cite{rombach2022high}, a robust LDM trained on large-scale datasets, which provides strong prior knowledge for generating high-quality garment. To capture garment features on a reference person, we first use IP-Adapter~\cite{ye2023ip} as a low-level feature extractor to get the coarse features of the garment, ensuring similarity between the generated garment and the reference. Additionally, we employ a high-level semantic extractor, called GarmNet, with the same structure as the Stable Diffusion’s denoiser, to capture detailed garment features from the person image. All the extracted features will be fused in the Garment Fusion (GarmFus) Blocks based on attention mechanism~\cite{vaswani2017attention}. Specifically, the low-level features are integrated into the denoiser through the cross-attention layers. For high-level semantic, we modify the self-attention layers to incorporate fine-grained features, ensuring the generated garment retains intricate details consistent with the reference image. 

Besides, the GarmRe and VITON are opposite tasks, their dataset has a certain degree of versatility. However, we notice that the dataset like VITON-HD~\cite{choi2021viton} typically used for VITON tasks cannot be directly applied to garment restoration task. As shown in \cref{fig:fig2_dataset}(a), there are three key reasons that can interfere with training. For the first row of example training pairs, regions with distinct garment characteristics are occluded or truncated, such as the lower half of a jumpsuit and the bottom part of trousers. Intuitively, the GarmRe model cannot infer whether the unseen region contain additional garment element, which can negatively affect training and result in extra garment generation in the final output. For the next two rows, however, cause the model to generate side view or extreme proportion garment image, which should be avoided. In contrast, we show suitable training data pairs in \cref{fig:fig2_dataset}(b), all of which share the characteristics of a front view and a suitable garment proportion in the image. For proper occlusions or truncations that do not obscure key garment features, we keep it as a hard case in the training dataset. Finally, we employ a coarse-to-fine training strategy to maximize the use of existing VITON datasets and filter out a GarmRe-specific dataset to make final fine-tuning. 

The structure and training strategy we employ ensure that our model generates high-quality restored garments as shown in \cref{fig:showresult}. Our contributions are summarized as follows:
\begin{itemize}
    \item   We propose IGR based on improved LDM to restore high-fidelity garment that is well-aligned with the reference person.
    \item	We introduce multiple extractors to fully capture both low-level features and high-level semantics of the garment on a person and propose the GarmFus to accurately integrate the extracted features into the generation process.
    \item	We implement a coarse-to-fine training strategy, starting with the existing VITON datasets to establish a foundation, and subsequently improving the generative performance with a dataset tailored for GarmRe.
\end{itemize}

\section{Related Work}
\label{sec:related}

\subsection{Conditional Diffusion Models}\label{sec:cDM}
Recently, the rapid advancement of latent diffusion models, led by Stable Diffusion~\cite{rombach2022high}, has driven significant progress in text-to-image generation~\cite{rombach2022high, podell2023sdxl, esser2024scaling, flux}. Several universal controllers~\cite{mou2024t2i,ye2023ip,zhang2023adding} have emerged to enhance control over generated images. IP-Adapter~\cite{ye2023ip} gives LDMs the ability to receive input image prompt as input to generate output containing their features, while attributes-specific ControlNet~\cite{zhang2023adding} enables generated results to align with given poses, outlines, depths, and other attributes. T2I-Adapter~\cite{mou2024t2i} introduces a unified model to accept various attributes. These controllers can be easily integrated into most downstream tasks~\cite{choi2024improving,xu2024magicanimate}. However, certain tasks require further specialization. For instance, pose-guided human generation~\cite{lu2024coarse,hu2024animate, xu2024magicanimate} demands additional controls to preserve identity during pose changes, and VITON~\cite{choi2024improving,kim2024stableviton,xu2024ootdiffusion} relies on extractors to capture complete features of a standard garment with clean background. GarmRe in this paper, as the inverse of VITON, necessitates more powerful extractors to derive comprehensive garment features as condition from reference person image.

\subsection{Garment Restoration}\label{sec:GR}
Garment Restoration aims to restore standard garment from person image. TileGAN~\cite{zeng2020TileGAN} is the first to propose a two-stage GAN-based~\cite{isola2017image, goodfellow2014generative} model for garment restoration: the first stage uses category guidance to generate a coarse garment rely on the person image, followed by refinement in the second stage. However, due to the limitations of its foundational model, the generated results are unsatisfactory. Recent works, such as TryOffDiff~\cite{velioglu2024tryoffdiff}, TryOffAnyone~\cite{xarchakos2024tryoffanyonetiledclothgeneration} and RAGDiffusion~\cite{tan2024ragdiffusion}, leverage pretrained LDM~\cite{rombach2022high} as their base models, achieving significant improvements due to their strong generative capabilities. TryOffDiff employs SigLIP~\cite{zhai2023sigmoid} to extract garment features from person images, while TryOffAnyone avoids additional structures by directly concatenating the person image into the model input. Despite these advancements, these two methods struggle to preserve fine garment details. RAGDiffusion can generate high-fidelity garments but introduces cumbersome process to achieve retrieval-augmented techniques like~\cite{borgeaud2022improving, ram2023context}. Our IGR improves the pretrained diffusion model and is able to preserve the garment details of person image and generate the authentic garment in a easy way.

\section{Method}
\label{sec:method}

\begin{figure*}[ht]
    \centering
    \includegraphics[width=0.9\linewidth]{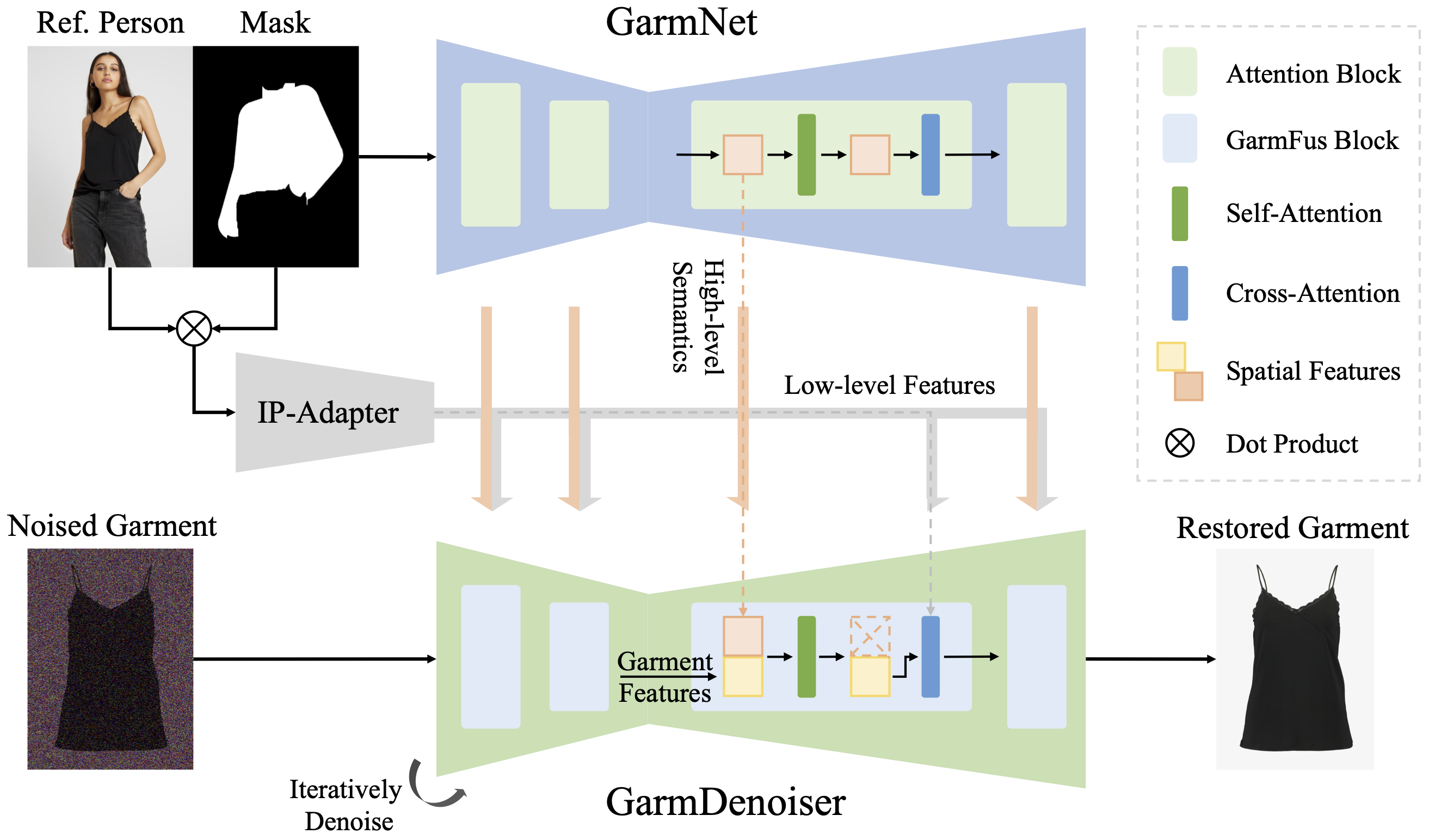}
    \caption{Overview of our proposed IGR. The GarmNet and GarmDenoiser share the same structural design. The reference person image, concatenated with the garment mask, serves as input to GarmNet, extracting high-level garment semantics. Simultaneously, the IP-Adapter captures low-level garment features from the segmented garment. The GarmDenoiser integrates garment features with intermediate outputs from GarmNet through self-attention layers, retaining only the latter half. Low-level features are incorporated via cross-attention layers for comprehensive garment reconstruction.}
    \label{fig:fig3_model}
\end{figure*}

\subsection{Preliminary: Latent Diffusion Models}\label{sec:LDM}
Our model is fine-tuned based on Stable Diffusion~\cite{rombach2022high}, one of the most widely used LDMs. To reduce computational load, it utilizes a VAE~\cite{kingma2013auto} encoder $\mathcal{E}$ to compress image $x$ into latent space, yielding $z=\mathcal{E}\left(x\right)$, which can then be reconstructed using the paired decoder. During training, the noisy latent $z_t$ is randomly sampled from the forward diffusion process of $z$ at a time step $t\sim\mathcal{U}[1,\mathrm{T}]$ with the added Gaussian noise $\epsilon\sim\mathcal{N}(0,\mathrm{I})$. And a denoiser $\epsilon_\theta$ with the UNet structure~\cite{ronneberger2015u}, composed of convolution layers, self-attention layers and cross-attention layers~\cite{vaswani2017attention}, is employed to predict noise $\epsilon$ conditioned on text embeddings $c$ encoded by CLIP text encoder~\cite{radford2021learning}. When only the denoiser requires training, the optimization objective function is:
\begin{equation}
\mathcal{L}_{\mathrm{LDM}}=\mathbb{E}_{\epsilon\sim\mathcal{N}(0,\mathrm{I}),t\sim\mathcal{U}[1,\mathrm{T}]}[\|\epsilon-\epsilon_{\theta}(z_{t};c,t)\|_{2}^{2}].
\end{equation}
To enhance classifier-free guidance~\cite{ho2022classifier}, our base model employs condition drop during training, enabling joint learning of unconditional and conditional noise prediction. During inference, the final predicted noise ${\widetilde{\epsilon}}_\theta$ is obtained by interpolating between the conditional and unconditional noise, achieving a balance between sample quality and diversity:
\begin{equation}
\tilde{\epsilon}_\theta(\mathrm{z}_\mathrm{t};c,t)=s\epsilon_\theta(\mathrm{z}_\mathrm{t};c,t)+(1-s)\epsilon_\theta(\mathrm{z}_\mathrm{t};t),
\end{equation}
where $s$ is a guidance scale that controls the strength of text condition $c$.

\subsection{Garment Restoration}\label{sec:GarmRe}
\noindent \textbf{Garment Extractor.} As illustrated in \cref{fig:fig3_model}, we use the person image $p\in\mathbb{R}^{3\times H\times W}$ and its agnostic mask $m\in\mathbb{R}^{1\times H\times W}$ which is commonly used in VITON task~\cite{choi2021viton, kim2024stableviton, cui2024street, choi2024improving, xu2024ootdiffusion} as inputs. To extract the garment features from the person image, we leverage IP-Adapter~\cite{ye2023ip} as the low-level feature extractor and GarmNet as the high-level semantic extractor. The IP-Adapter allows our base model to take person image as prompt by utilizing the CLIP Image Encoder~\cite{radford2021learning}. Since the encoder input resolution is limited to $224\times224$, we first crop the garment area of the person image with paired agnostic mask and then resize it to the target resolution to retain as many garment details as possible, as $p_g=\text{Resize}\left(\text{Crop}\left(p\cdot m\right)\right)$. The cropped garment $p_g$ is input into the encoder and feature projection layers to extract the low-level garment features $F_{ll}$.

Relying solely on IP-Adapter, our garment restoration model struggles to achieve high consistency with the garment on reference person, often leading to detail mismatches. To address this, we introduce a high-level semantic extractor called GarmNet, which shares a same structure with the GarmDenoiser. We don't directly use the cropped garment as input. Instead, we concatenate the channel dimension of the person image latent $\mathcal{E}\left(p\right)\in\mathbb{R}^{4\times \frac{H}{8}\times \frac{W}{8}}$ and the resized mask $m_{resized}\in\mathbb{R}^{1\times \frac{H}{8}\times \frac{W}{8}}$ to get the GarmNet input with 5 channels. This approach avoids potential inaccuracies in the mask and prevents the loss of garment features caused by direct cropping. By learning to utilize the mask effectively, GarmNet captures robust garment characteristics while mitigating errors from incorrect masking. The intermediate features extracted at different resolutions before the self-attention layers are then used as high-level garment semantics $F_{hl}$.

\noindent \textbf{GarmDenoiser.} To integrate the low-level features $F_{ll}$ and high-level semantics $F_{hl}$ extracted from the reference person image into the denoising process, we employ GarmFus blocks based on attention mechanisms~\cite{vaswani2017attention}. 
For low-level garment features extracted by IP-Adapter, an additional cross-attention layer is added alongside the original cross-attention layer in the Stable Diffusion. Then, we get the low-level fused output as follows:
\begin{equation}
\text{Attn}(Q_g, K_{c}, V_{c}) + \text{Attn}(Q_g, K_{ll}, V_{ll}),
\end{equation}
where, $Q_g$ is derived from the intermediate features of the GarmDenoiser, while $K_{c}$ and $V_{c}$ are obtained from the text condition $c$, and $K_{ll}$ and $V_{ll}$ are sourced from the low-level garment features $F_{ll}$. $\text{Attn}$ is the calculation mode from \cite{vaswani2017attention}.
For high-level semantics, we modify the self-attention layers in the denoiser following the approach proposed in \cite{hu2024animate}. As illustrated in the bottom half of \cref{fig:fig3_model},  garment features $F_g\in\mathbb{R}^{c\times h\times w}$ are concatenated with the corresponding $F_{hl}\in\mathbb{R}^{c\times h\times w}$ along the spatial dimension to get the $F_{con} = \text{Concat}\left(F_{hl}, F_g\right)\in\mathbb{R}^{c\times h\times 2w}$ and passed through self-attention layers.  Then, we get the high-level fused output as follows:
\begin{equation}
\text{Attn}(Q_{fcon}, K_{fcon}, V_{fcon}),
\end{equation}
where, $Q_{fcon}$, $Q_{fcon}$ and $V_{fcon}$ are all sourced from $F_{con}$. Then, only the latter half with dimension of ${c\times h\times w}$ is retaining as the output.
These feature integration strategies enable our IGR to generate garments that are highly realistic and closely aligned with those in the reference person image.
For the text condition, we use three different prompts corresponding to categories: upper, lower and full body.

\noindent \textbf{Training Strategy.} We observe that datasets typically used for VITON tasks are not directly applicable to the GarmRe task. To maximize the utility of existing datasets while addressing the data issues highlighted in \cref{fig:fig2_dataset}(a), we adopt a coarse-to-fine training strategy. In the coarse training stage, IGR is initially trained on the VITON dataset, leveraging its extensive and diverse data pairs to establish a robust base model. For the fine-tuning stage, we curate a high-quality subset from the coarse dataset by manually removing garments with defects such as inappropriate occlusions, truncations, and side views. Additionally, images with extreme proportions are resized and padded to achieve suitable proportions. This GarmRe-specific dataset, consisting of training pairs shown in \cref{fig:fig2_dataset}(b), is then used to fine-tune the model for improved performance.

\section{Experiments}
\label{sec:experiment}

\subsection{Experimental Setup}\label{sec:exp_set}
\noindent \textbf{Datasets.} 
Our proposed IGR is initially trained on a coarse dataset consisting of the VITON-HD~\cite{choi2021viton} and a self-collected paired dataset of in-shop models and garments. Then, we selected 20\% of the data pairs from the coarse dataset for the further fine-tuning. As with other works~\cite{xu2024ootdiffusion,choi2024improving,velioglu2024tryoffdiff}, we divide VITON-HD into a training dataset and a testing dataset. The majority of images in this dataset feature person image facing forward against clean background, with tiny occlusion of the garment. To more comprehensively analyze our method's performance, we also adopt the testing person images in StreetTryOn~\cite{cui2024street} dataset.

\noindent \textbf{Implementation Details.} 
We adopt the Stable Diffusion~\cite{rombach2022high}(v1.5) as our foundation model. The AdamW optimizer~\cite{loshchilov2017decoupled} with hyper-parameters set to a batch size of 32, a learning rate of 1e-5 is employed to train the models at a resolution of $1024\times768$. All experiments are conducted on four NVIDIA A100 80 GB GPUs with DeepSpeed ZeRO-2~\cite{rajbhandari2020zero} to save memory usage. At inference stage, we run IGR for 25 sample steps with the DDIM sampler~\cite{song2020denoising} to get the final garment.

\subsection{Qualitative and Quantitative Comparison}\label{sec:exp_comp}
\begin{figure*}[ht]
    \centering
    \includegraphics[width=0.95\linewidth]{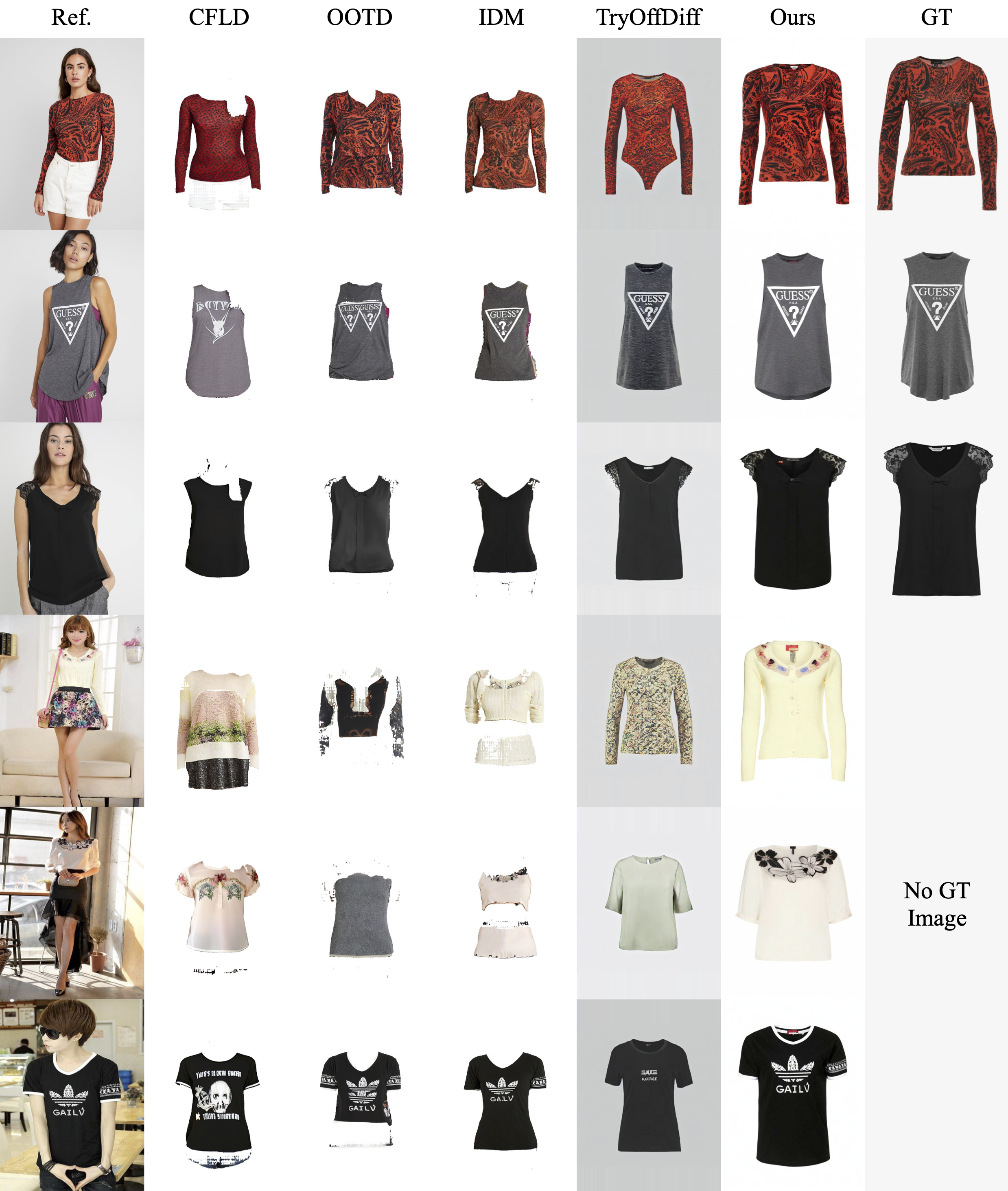}
    \caption{Qualitative comparison. The first three rows are tested on the VITON-HD dataset, while the remaining rows are tested on the StreetTryOn dataset. GT refers the ground truth.}
    \label{fig:fig4_method}
\end{figure*}

\begin{table*}[htbp]
\centering
\begin{tabular}{l|cccccc|ccc}
\toprule
\multirow{2}{*}{Model} & \multicolumn{6}{c|}{Paired Test on VITON-HD}            & \multicolumn{3}{c}{Unpaired Test on StreetTryOn} \\ \cmidrule(){2-10} 
                       & SSIM$\uparrow$    & LPIPS$\downarrow$  & DISTS$\downarrow$  & FID$\downarrow$     & CLIP-FID$\downarrow$ & KID*$\downarrow$    & FID$\downarrow$             & CLIP-FID$\downarrow$       & KID*$\downarrow$          \\ \midrule
CFLD                   & 0.7489 & 0.4320 & 0.2979 & 46.2532 & 10.3878  & 33.8882 & 62.1068         & 12.8040        & 35.7497       \\
OOTD                   & 0.7219 & 0.4424 & 0.2984 & 66.4205 & 10.0482  & 52.3561 & 113.5400        & 16.4324        & 89.4453       \\
IDM                    & 0.7402 & 0.4374 & 0.2939 & 63.0603 & 11.1035  & 50.9523 & 103.1404        & 16.3456        & 76.8308       \\ \midrule
TryOffDiff             & 0.7576 & 0.4288 & 0.2680 & 28.8638 & 14.7685  & 16.4532 & 39.8455         & 19.7795        & 20.5915       \\
Ours                   & \textbf{0.7895} & \textbf{0.2946} & \textbf{0.2045} & \textbf{13.1425} & \textbf{5.0746}   & \textbf{2.9724}  & \textbf{31.0537}         & \textbf{10.0137}        & \textbf{8.7535}       \\ \bottomrule
\end{tabular}
\caption{Quantitative comparison for baseline models and ours. The KID metric is multiplied by the factor 1e3 to ensure a similar order of magnitude to the other metrics.}
\label{tab:quantitative_method}
\end{table*}

For the baselines, we extended the settings of TryOffDiff~\cite{velioglu2024tryoffdiff} and adopt several methods to achieve garment restoration. For CFLD~\cite{lu2024coarse}, a straight pose is used to adjust the attitude of the reference person, while for OOTD~\cite{xu2024ootdiffusion} and IDM~\cite{choi2024improving}, frontal fitting model is employed to try on cropped garment from reference person. These configurations ensure that the generated person image with the garment face forward, avoiding unnecessary occlusions. The garment is then segmented directly from person image using SAM2~\cite{ravi2024sam}. We also compare our model with the tailored TryOffDiff which can directly output the restored garment. To comprehensively evaluate the performance of our proposed IGR model, we conduct both qualitative and quantitative comparisons of the generated garment against real garment on the VITON-HD and StreetTryOn testing dataset.

\noindent \textbf{Qualitative Comparison.} \cref{fig:fig4_method} presents a qualitative comparison between our proposed IGR and baseline models on two testing datasets. While CFLD~\cite{lu2024coarse}, OOTD~\cite{xu2024ootdiffusion} and IDM~\cite{choi2024improving} are not specifically designed for GarmRe task, they can straighten garment but fail to retain detailed features aligned with the reference person. Additionally, segmentation model introduces additional artifacts in the generated garment. TryOffDiff is tailored for GarmRe and trained on the VITON-HD dataset. It performs well on the VITON-HD testing dataset. However, as shown in the first row of \cref{fig:fig4_method}, TryOffDiff generates extra garments due to direct training on the VITON dataset, as previously discussed. Furthermore, it struggles to preserve garment details when tested on the StreetTryOn dataset, where the garment on person is more complex. In contrast, our IGR consistently excels at preserving garment details, regardless of the complexity of the person image, generating high-fidelity garment. To further validate the generalization capabilities of IGR, we test it on cosplay and unreal person images, which differ significantly from the training data. Despite these differences, our model generates plausible garments, as partially shown in \cref{fig:showresult}, by leveraging the strong prior knowledge embedded in the foundational model. Intuitively, improvements in the foundational model's capabilities will lead to even higher-quality garment restoration.

\noindent \textbf{Quantitative Comparison.} As shown in \cref{tab:quantitative_method}, we evaluate the fidelity of the generated garment distributions for our IGR model and baseline models on two testing datasets using FID~\cite{heusel2017gans}, CLIP-FID~\cite{kynkaanniemi2022role} and KID~\cite{binkowski2018demystifying} metrics. Since StreetTryOn lacks real garments compared to VITON-HD, the real garments from VITON-HD are shared for calculating these metrics. Additionally, for testing results on the VITON-HD dataset, we assessed three more metrics: SSIM~\cite{wang2004image}, LPIPS~\cite{zhang2018unreasonable} and DISTS~\cite{ding2020image} to evaluate the reconstruction quality between the generated garment and corresponding ground truth (GT). The experimental results indicate that the pose-transfer-based CFLD achieves more stable and realistic garments compared to the Try-on-based methods OOTD and IDM. Although the tailored TryOffDiff has made great progress, our model demonstrates more superior performance in quantitative comparisons. This is precisely because we adopt stronger feature extractors and better training strategies.

\subsection{Ablation Study}\label{sec:exp_ablation}
\begin{figure}[htb]
    \centering
    \includegraphics[width=0.95\linewidth]{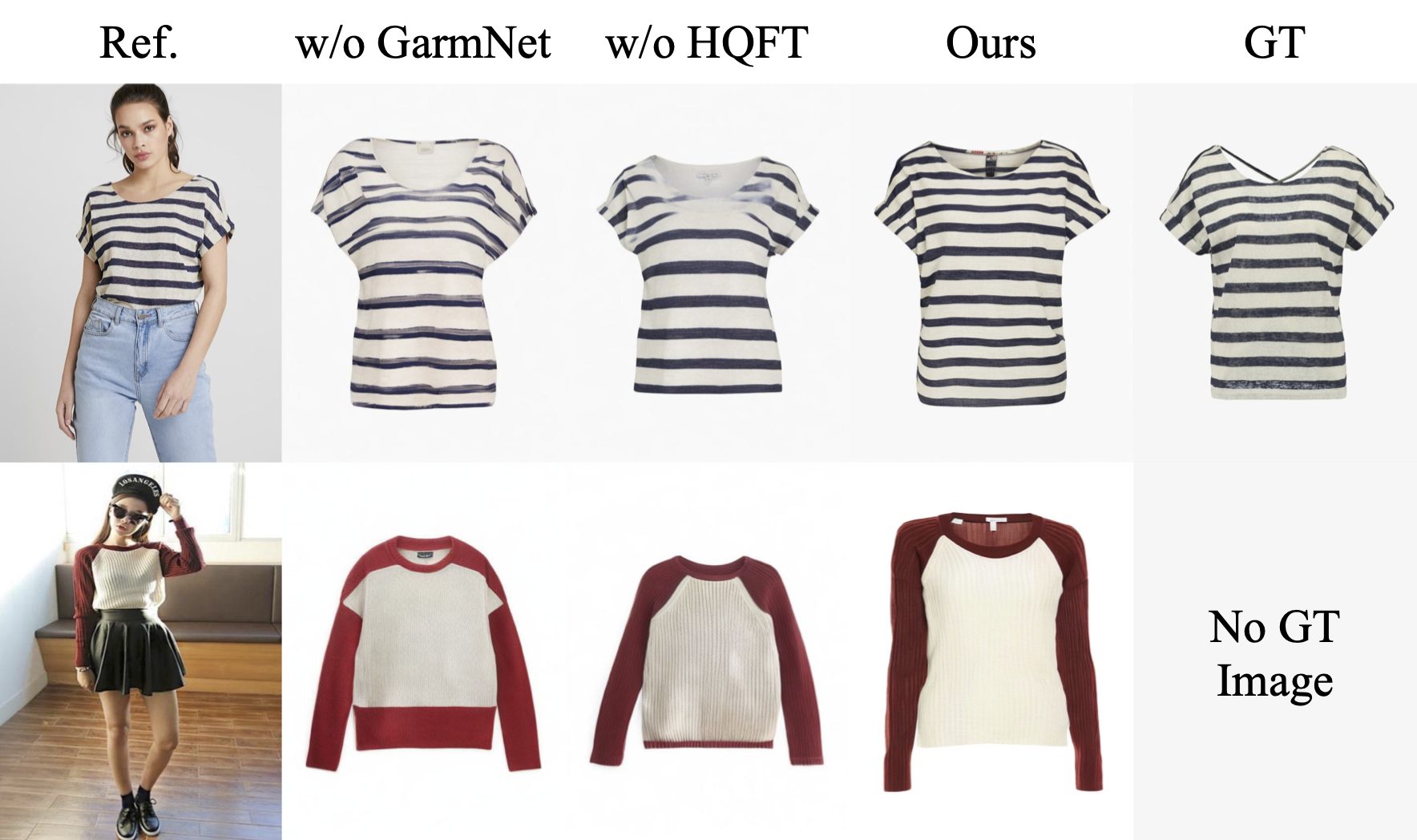}
    \caption{Qualitative ablations for structure and training strategy.}
    \label{fig:fig5_structure}
\end{figure}

\begin{table*}[htbp]
\centering
\begin{tabular}{l|cccccc|ccc}
\toprule
\multirow{2}{*}{Model} & \multicolumn{6}{c|}{Paired Test on VITON-HD}            & \multicolumn{3}{c}{Unpaired Test on StreetTryOn} \\ \cmidrule(){2-10} 
                       & SSIM$\uparrow$    & LPIPS$\downarrow$  & DISTS$\downarrow$  & FID$\downarrow$     & CLIP-FID$\downarrow$ & KID*$\downarrow$    & FID$\downarrow$             & CLIP-FID$\downarrow$       & KID*$\downarrow$          \\ \midrule
w/o GarmNet                  &  0.7931 & 0.3048 & 0.2192 & 15.7038 & 5.6490 & 4.4113 & 44.4640 & 12.1343 & 18.4922 \\

w/o HQFT                   & \textbf{0.7984} & 0.2988 & 0.2054 & 13.6753 & 5.9490 & 3.5486 & 41.0889 & 11.1313 & 15.9371 \\
Ours                   & 0.7895 & \textbf{0.2946} & \textbf{0.2045} & \textbf{13.1425} & \textbf{5.0746}   & \textbf{2.9724}  & \textbf{31.0537}         & \textbf{10.0137}        & \textbf{8.7535}       \\ \bottomrule
\end{tabular}
\caption{Quantitative ablations for structure and training strategy. The KID metric is multiplied by the factor 1e3 to ensure a similar order of magnitude to the other metrics.}
\label{tab:quantitative_structure}
\end{table*}

\noindent \textbf{Effect of structure.} 
We perform an ablation study to evaluate the role of GarmNet in preserving the fine-grained details of target garment on a person. For comparison, we train an another model that excludes GarmNet, relying solely on the IP-Adapter~\cite{ye2023ip} to extract garment features from the person image. The qualitative results are presented in \cref{fig:fig5_structure}. While training with only the IP-Adapter can reconstruct garment similar to the garment on reference person, many fine details are missing. In contrast, incorporating GarmNet enables better alignment of texture details with the reference person image. As shown by the quantitative metrics in \cref{tab:quantitative_structure}, GarmNet effectively captures garment details, achieving higher scores in both reconstruction accuracy and authenticity.

\noindent \textbf{Effect of training strategy.} We notice differences in the training dataset between VITON and GarmRe and adopted a training strategy from coarse to fine. With our proposed high-quality fine-tuning (HQFT), the generated garment becomes more realistic, without excessive garment details, as shown in the bottom of \cref{fig:fig5_structure}. The training strategy we proposed further enhances the fidelity of the generated garment as shown in the last row of \cref{tab:quantitative_structure}.

\begin{figure}[htb]
    \centering
    \includegraphics[width=0.98\linewidth]{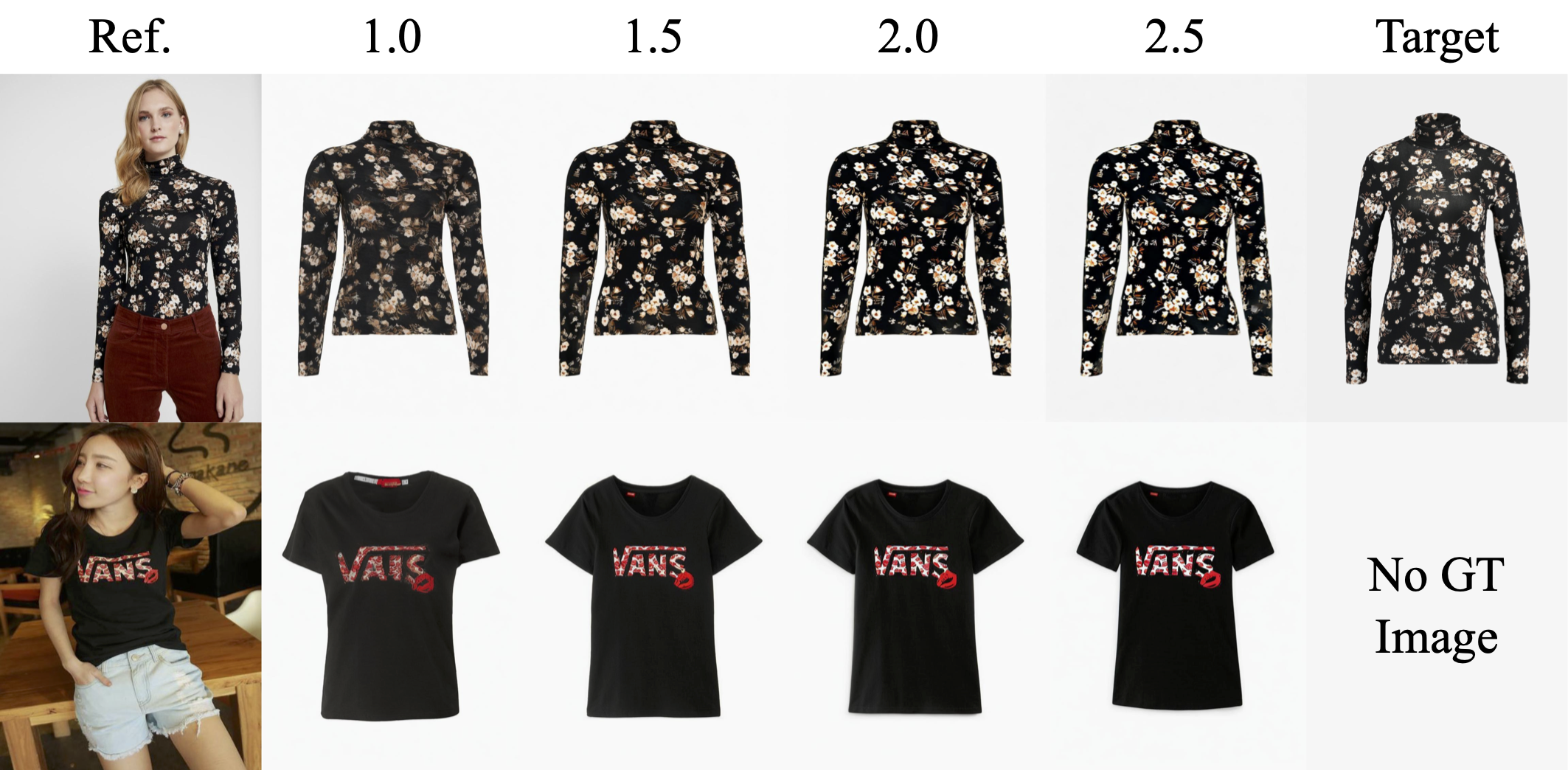}
    \caption{Qualitative ablations for guidance scale.}
    \label{fig:fig6_cfg}
    \vspace{-2mm}
\end{figure}
\begin{table*}[htbp]
\centering
\begin{tabular}{l|cccccc|ccc}
\toprule
Guidance & \multicolumn{6}{c|}{Paired Test on VITON-HD}            & \multicolumn{3}{c}{Unpaired Test on StreetTryOn} \\ \cmidrule(){2-10} 
                      Scale  & SSIM$\uparrow$    & LPIPS$\downarrow$  & DISTS$\downarrow$  & FID$\downarrow$     & CLIP-FID$\downarrow$ & KID*$\downarrow$    & FID$\downarrow$             & CLIP-FID$\downarrow$       & KID*$\downarrow$          \\ \midrule
1.0 & \textbf{0.7940} & \underline{0.2957} & 0.2110 & 15.8122 & 5.5150 & 4.7267 & 31.8787 & \textbf{9.6373}  & 9.5388 \\
1.5 & \underline{0.7895} & \textbf{0.2946} & \underline{0.2045} & 13.1425 & \underline{5.0746} & 2.9724 & \underline{31.0537} & \underline{10.0137} & \textbf{8.7535} \\
2.0 & 0.7781 & 0.2961 & \textbf{0.2043} & \textbf{12.2856} & \textbf{5.0132} & \underline{2.4130} & \textbf{30.8504} & 10.3480 & 8.8750 \\
2.5 & 0.7651 & 0.2979 & 0.2059 & 12.0341 & 5.1168 & \textbf{2.2296} & 31.4109 & 10.7149 & \underline{8.7850} \\ \bottomrule
\end{tabular}
\caption{Quantitative ablations for guidance scale. The KID metric is multiplied by the factor 1e3 to ensure a similar order of magnitude to the other metrics.}
\label{tab:quantitative_cfg}
\end{table*}

\noindent \textbf{Effect of guidance scale.} As our model is trained with condition drop, we set the guidance scale to 1.0, 1.5, 2, 2.5, 3.0 for classifier-free guidance. Quantitative results shown in \cref{tab:quantitative_cfg} indicate that guidance scales of 1.5 and 2.0 perform similarly on the VITON-HD dataset. However, for the StreetTryOn dataset, a guidance scale of 1.5 is preferable. This is because the StreetTryOn dataset features more complex person images, including significant garment occlusions and distortions. In such cases,  the model's inherent capability to restore the garment plays a more critical role than relying on a higher guidance scale, which increases dependency on the extracted garment features. As shown in \cref{fig:fig6_cfg}, a guidance scale of 1.5 produces visually pleasing results on both testing datasets. Lower guidance scales result in blurry garments with a loss of detail, while higher scales lead to overly saturated garments.
\section{Conclusion}
\label{sec:conclusion}

In this paper, we present IGR, a model improved LDM to restore standard garments from person images, which is the inverse of the VITON task. Our approach leverages Stable Diffusion as the foundation denoiser, offering robust generalization capabilities. And two separate feature extractors: IP-Adapter and GarmNet are employed to capture low-level features and high-level semantics of the garment on person, respectively. To ensure that the generated garment closely aligned with the garments in the person images, we introduce GarmFus blocks with attention mechanism to fully fuse the extracted garment features in the GarmDenoiser. Furthermore, we observe the differences in the training data between GarmRe and VITON, prompting the adoption of a coarse-to-fine training strategy to fully utilize the VITON dataset and improve garment fidelity. Extensive experiments show that our proposed IGR surpasses existing methods, producing garment with remarkable realism and intricate detail.

{\small
\bibliographystyle{ieeenat_fullname}
\bibliography{11_references}
}


\end{document}